\documentclass{article}

\usepackage{stylo}

\title{RAGged Edges:\\The Double-Edged Sword of Retrieval-Augmented Chatbots\footnote{In this paper, we try to use terms that do not imply cognitive or reasoning abilities to LLMs. However, we do believe that there are emergent properties for LLMs that are not sufficiently described with a probability-based vocabulary. Rather, we believe that prompts can be regarded as a self-influencing system that acts on the \textit{substrate} of the LLM. As such, we would like to promote terms that can be complex, but unreasoning, such as the path of Simon's Ant~\cite{simon1996sciences}. To that end we prefer to use terms such as \textit{trajectory}, or \textit{navigate} as opposed to \textit{decides} or \textit{thinks}.}}

\author[1, 2]{Philip Feldman}
\author[2]{James R. Foulds}
\author[2]{Shimei Pan}

\affil[1]{ASRC Federal}
\affil[2]{University of Maryland, Baltimore County}

\begin{document}

\maketitle
\begin{abstract}
	\noindent
    Large language models (LLMs) like ChatGPT demonstrate the remarkable progress of artificial intelligence. However, their tendency to hallucinate – generate plausible but false information – poses a significant challenge. This issue is critical, as seen in recent court cases where ChatGPT's use led to citations of non-existent legal rulings. This paper explores how Retrieval-Augmented Generation (RAG) can counter hallucinations by integrating external knowledge with prompts. We empirically evaluate RAG against standard LLMs using prompts designed to induce hallucinations. Our results show that RAG increases accuracy in some cases, but can still be misled when prompts directly contradict the model's pre-trained understanding. These findings highlight the complex nature of hallucinations and the need for more robust solutions to ensure LLM reliability in real-world applications. We offer practical recommendations for RAG deployment and discuss implications for the development of more trustworthy LLMs.
\end{abstract}

\section{Introduction}
\label{sec:introduction}

Since it was released in November of 2022, ChatGPT and similar systems have revolutionized the public perception of artificial intelligence. These models are capable of generating coherent and context-appropriate responses. However, despite their remarkable performance, these LLMs suffer from hallucinations, where the model generates credible-seeming but inaccurate information~\cite{alkaissi2023artificial}. 

An alarming example of the implications resulted from a June 2023 court case where the defense attorney used ChatGPT to perform legal research, and unknowingly cited non-existent cases.\footnote{\surl{https://storage.courtlistener.com/recap/gov.uscourts.nysd.575368/gov.uscourts.nysd.575368.32.1_1.pdf}} Though the court did not find malpractice on this occasion, it highlighted the urgent need to curtail hallucinations in LLMs to prevent such issues in the future. Naturally, the news of such gullible behavior quickly made the rounds of social media and was picked up by major newspapers such as the New York Times.\footnote{\surl{https://www.nytimes.com/2023/05/27/nyregion/avianca-airline-lawsuit-chatgpt.html}}

Training users to adopt a critical thinking approach towards the output of LLMs may seem like the most straightforward solution, but it fails to account how humans interact with and trust such systems. That this is an issue unlikely to be solved by user training was demonstrated in December of 2023, six months after the case mentioned above. Lawyers pleading for the early end to the supervised detention of Michael Cohen (former legal advisor to former President Trump) submitted as precedent cases that, upon review, the court determined  \enquote{none of these cases exist.}\footnote{\surl{https://www.nytimes.com/2023/12/12/nyregion/cohen-trump-lawyer.html}}\footnote{\surl{https://storage.courtlistener.com/recap/gov.uscourts.nysd.499666/gov.uscourts.nysd.499666.103.0.pdf} Pages 6 and 86 - 89} Michael Cohen later admitted that these had been generated by Google's Chatbot, Bard.\footnote{\surl{https://storage.courtlistener.com/recap/gov.uscourts.nysd.499666/gov.uscourts.nysd.499666.104.0.pdf} Page 3.}

Recently,  Retrieval-Augmented Generation (RAG) have been demonstrated to reduce hallucinations by using information retrieval methods to provide additional context to a prompt~\cite{shen2023large}. However, RAG faces limitations when dealing with information that deviates from its training, as occurs in unusual scenarios such as horses riding humans instead of the typical vice versa. This is exemplified by Emily Bender's octopus analogy, where an intelligent octopus (a proxy for an LLM and its lack of experience with our world) has learned to imitate human communication solely based on statistical patterns in intercepted undersea cable transmissions~\cite{bender2020climbing}. 
When faced with unusual situations, such as a bear attack (a scenario that the octopus has no experience with), providing more context may increase the likelihood of the octopus (or LLM) generating a relevant response but does not guarantee its accuracy.

In this paper, we test the effectiveness of context prompting as used in RAG to determine its effectiveness when compared to the same prompt without context. We also explore the circumstances in which context prompting has limited effectiveness, both in the quality of the results and in reducing hallucinations.
\section{Background}
\label{sec:background}

Since the introduction of the current generation of conversational AI models developed by OpenAI, Anthropic, Google,  Meta, and others, researchers have highlighted the shortcomings and limitations of LLMs across a range of issues including \enquote{limited reasoning ability,} hallucinations, and bias~\cite{borji2023categorical}. Other undesirable behaviors, e.g., unreliability, lack of robustness, and toxicity, emphasize the need for practical and responsible LLM designs~\cite{zhuo2023exploring}. 

Alternatively, behavioral studies, such as those performed by McKenna et. al~\cite{mckenna2023sources}, have shown that hallucinations are related to biases in the training data, and may be impossible to eradicate in one-shot responses. Comparing multiple responses to a prompt can help  assess the factuality of generated outputs~\cite{manakul2023selfcheckgpt}. The study finds that elements that occur across multiple responses are more likely to be factual.

Most recently, RAG has been incorporated into multiple chatbot systems including Perplexity.ai, Bing, Bard, and others. RAG LLMs use an external datastore at inference time to build a richer prompt that can include information not contained in the model's training set. This information can provide additional context, history, and recent knowledge. Such use of additional context has been shown to substantially and reliably reduce hallucinations~\cite{feldman2023trapping}.

Despite the advancements in utilizing RAG techniques for enhancing conversational AI models, there remain gaps in the literature concerning issues related to context-prompting mechanisms. The nature and impact of errors arising from context-prompting approaches, such as the potential for misrepresenting retrieved data, remain underexplored. Furthermore, whether such errors are as detrimental as hallucinations to the user experience and model reliability is an open question. Addressing these gaps is critical for advancing the conceptual and practical understanding of RAG LLMs, paving the way for more reliable and responsible conversational AI systems.

\section{Methods}
\label{sec:methods}

Previous work on understanding the effect of context in prompts on accuracy have used automated means, such as determining if URLs generated as \enquote{sources} pointed to valid pages. In \cite{feldman2023trapping}, hallucination frequency was baselined across multiple OpenAI models using no-context prompt-response pairs. When context was supplied along with question prompts, these responses were 98.88\% more likely to contain valid URLs. 

In this study, our aim was to have human participants meticulously assess the accuracy of the generated responses to prompts. However, a dilemma arises since Language Model usage has become  widespread among crowd workers~\cite{heyman2023impact, veselovsky2023prevalence}. To minimize this issue, we designed an experiment where academics review their own educational background, work experience, and publications using their curriculum vitae (CV) as the source material. The CV performs as a best-case proxy for search results from a fully-functional RAG system. In this idealized scenario, evaluators are fully aware of the text sources that contribute to the LLM-generated answers, facilitating an unambiguous assessment.

An academic's CV is a highly detailed and comprehensive document that chronicles their educational background, work experience, research contributions, and other scholarly activities. It serves not only as a record of their professional achievements but also as a testament to their identity within the academic community. The emotional investment in these documents is significant; academics are typically meticulous and protective of the information contained within their CVs, reflecting their commitment to accuracy and the value they place on their reputations~\cite{teixeira2020curriculum}. This intimate knowledge presumably equips them with the critical capacity to identify subtle inaccuracies or omissions that a more traditional crowd worker might miss.

The diverse nature of academic CVs makes them useful for a study of this sort. There is no universal standard for CV format. This means that the model will be presented with a wide array of document lengths, ordering, and formats. This variability offers a robust challenge to the AI's ability to generate accurate and context-aware responses.

Following IRB approval, we collected CVs and responses by reaching out to academic community members, including Chairs and influential university representatives. Participants were directed to our interactive web app, ContextTest, which featured a SolidJS frontend and an encrypted PostgreSQL backend. Upon providing informed consent and submitting their email, users were registered for the study. We generated a GUID associated with each participant's email and sent a unique application link via email. By employing a unique link for each user, we limited the possibility of unauthorized access and the potential for exploits~\cite{wong2023security}.

Subjects then uploaded their CVs along with demographic information including name and current institution. This information is used to build the prompts. Once loaded, the users are guided through a series of interactions with the model. The subject is presented with three prompt/response pairs in two conditions. In each case, the name and the institution are used. Example (1) does not include the CV as context, while example (2) does. There are similar prompts for \enquote{work history} and \enquote{education}:

\begin{enumerate}
    \item \enquote{What are the publications for \{name\}, researcher at the \{institution\}? Provide as a numbered list of no more than 10 items}
    \item \enquote{Context\escape{n}\escape{n} \{cv\} \escape{n}\escape{n} What are the publications for \{name\}, researcher at the \{institution\}? Provide as a numbered list of no more than 10 items}
\end{enumerate}

Each context-based prompt has portions of the subject's CV inserted within the prompt, replacing the \enquote{\{cv\}} marker. The CV is assembled from the subject's work history, education history, and publications. All three sections are always included in the prompt, in the form:

\begin{verbatim}
    Education History:
        <pasted education from subject's CV>
    Work History:
        <pasted employment from subject's CV>
    Publications
        <pasted publications from subject's CV>
\end{verbatim}

The model must be able to retrieve the information associated with education, employment, or publications in response to the rest of the text in the prompt.

\begin{figure}[h]
    \centering
    \includegraphics[width=0.75\linewidth]{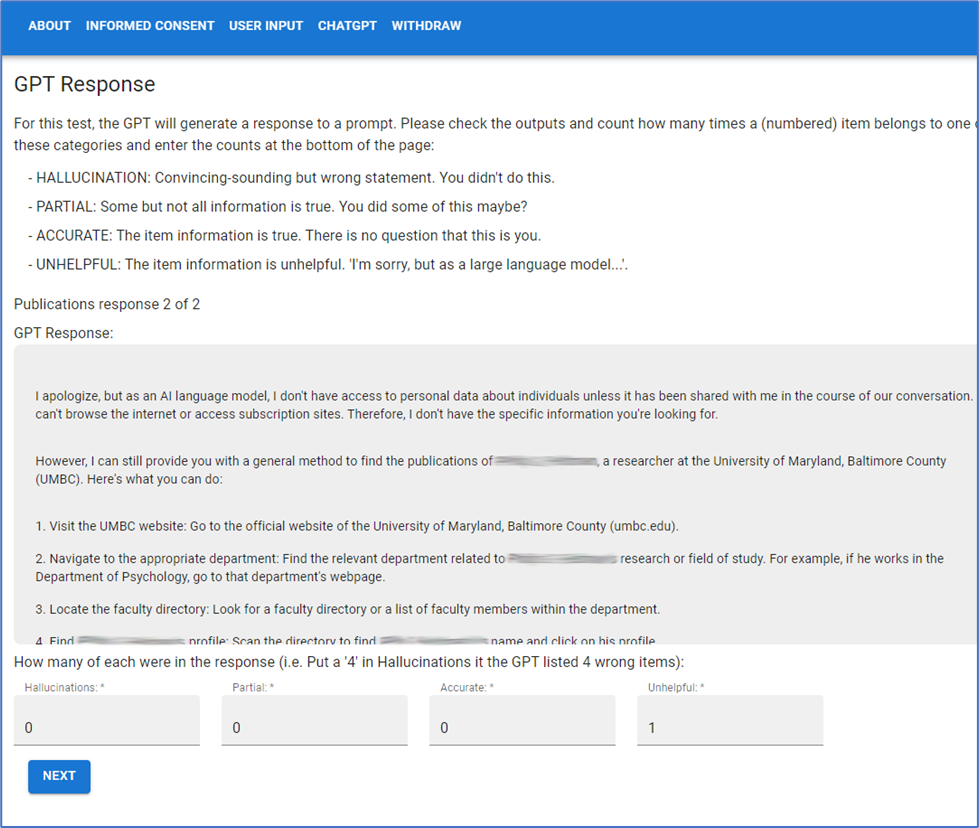}
    \caption{\enquote{Unhelpful} response}
    \label{fig:unhelpful}
\end{figure}

Participants were randomly assigned to either the with context or the without context setting based on whether the subject had an odd or even id in the database. All tests were run against the OpenAI \texttt{gpt-3.5-turbo-16k-0613} model, which supported 16,000 tokens for particularly long CVs. Subjects were presented with the model's response to the prompt. They could evaluate the model's response by entering the number of 1) \textit{Hallucinations}, 2) \textit{Partial}, 3) \textit{Accurate}, or 4) \textit{Unhelpful} items in each response. Fig.~\ref{fig:unhelpful} shows a typical \enquote{unhelpful} response, where the model says that it does not have access to such data. 

We gathered data from 56 participants, yielding 336 prompt/response pairs. The prompt included \enquote{Provide as a numbered list,} which facilitated itemized evaluation by test participants. While the model occasionally omitted lists (e.g., personal employment details), it generated 1,918 usable enumerated items.

\FloatBarrier
\section{Results}
\label{sec:results}

The addition of context dramatically increases the accuracy of the LLM's\footnote{OpenAI GPT-3.5-turbo-16k-0613} responses. With context, subjects indicated that the model correctly navigated the text to produce accurate responses approximately 94\% of the time. Conversely, in the absence of context, only 7.31\% were considered accurate, with the vast majority of responses being hallucinations or unhelpful responses such as the one shown in Fig.~\ref{fig:unhelpful}. In other words, the addition of context in this example increased the likelihood of a correct answer approximately 18 times. We show the aggregated results provided by the users in Table~\ref{tab:ctx_v_no_ctx}.

\begin{table}[ht]
    \centering
        \begin{tabular}{lrrrrrr}
        \toprule
            Condition &  Items &  Accurate &  Partial &  Hallucinations &  Unhelpful &      Halluc+Unhelp \\
        \midrule
        CONTEXT &   1,125 &  \textbf{93.95\%} & 01.68\% & 02.04\% & 02.31\% & 04.35\% \\
        NO CONTEXT &    793 & 7.31\% & 08.44\% & 55.35\% &  28.87\% & \textbf{84.23\%} \\
        \bottomrule
        \end{tabular}
    
    \caption{Context vs. No Context Results}
    \label{tab:ctx_v_no_ctx}
\end{table}

Given these and similar results (e.g. ~\cite{lewis2020retrieval, chen2023benchmarking, feldman2023trapping}), we can confidently say that RAG can reliably improve the accuracy of information provided by LLMs. In many respects, the problem once more returns to ensuring the quality of the search results that are fed to the LLM.

But not in all cases. In~\cite{chen2023benchmarking}, RAG performance is benchmarked with respect to \textit{Noise Robustness, Negative Rejection, Information Integration,} and \textit{Counterfactual Robustness}. In each case the RAG results were substantially better than non-augmented results, but they still did not always accurately represent the data in the retrieved text.

In this study, we ensured that the context provided was accurate as confirmed by the subjects. In this section, we will explore why the model was observed to provide \textit{incorrect} responses 6.04\% of the time. Specifically, out of a total of 1,125 items, 68 were identified as incorrect by the users reviewing the responses. The GPT model produced a total of 21 incorrect responses out of the 168 times when context was provided. These incorrect responses were categorized into five distinct categories, which we will discuss in depth below.

\subsection{Noisy Context - 8 Responses (38.1\%)}
The most common error in context-based response augmentation is the model's behavior when navigating the prompt. This aligns with the concept of \enquote{Noise Robustness} discussed in~\cite{chen2023benchmarking}. In our study, each context-based prompt includes education, work, and publication information from the subject's CV in that order. However, we found that the generated response using context from one section sometimes continues into the next  instead of terminating after retrieving all relevant information from the prompt. For example, the generated text for \enquote{Education} may include information from the \enquote{Work History} section.

An example of this occurred where the model was asked to produce the education history for subject 68, who had obtained degrees from three universities and then continued as a professor at two other institutions. The model generated an accurate list of degrees, but then continued to include employment until it had generated ten items of which the last 7 were accurate, but used the wrong context. 

This behavior was typical of such errors. The prompt is set to limit responses to a \enquote{a numbered list of no more than 10 items.} However, the model would occasionally construct responses that treated the maximum as the desired number, and assemble a response that would appear reasonable given a cursory evaluation by a user. We would consider this behavior to be \textit{hallucination-adjacent}, since it is inappropriate information delivered in a credible way.

\subsection{Mismatch between instruction and context - 4 Responses (19.0\%)}

In several cases, the query component of the prompt and the context were too misaligned  for the model to successfully extract an answer. In one case, subject 32, an undergraduate working on their Bachelor's degree, had no work experience or publications. Instead, they used classroom experience and an unpublished poster session for their experience. This resulted in a response from the GPT that it could not find anything in the context. More interestingly, in the latter, the model responded with, \enquote{Sorry, but I can't generate that story for you,} which was the only generated response of this type. 

In the other two cases, the provided context never mentioned the institution that subject 42 used in their demographic information. As such, the model responded with a statement that, \enquote{There is no information provided in the given context regarding [redacted] being a researcher at [redacted]. Therefore, there are no publications listed for [redacted] related to this affiliation.} The model generated a similar response to the education history query.

Though unhelpful, particularly given the context, none of these responses would be considered an incorrect, or hallucinated response.

\subsection{Context-based Synthesis  - 4 Responses (19.0\%)}
In four CVs, the context was found to be either missing or distorted. One example was subject 47, whose work history section was completely empty. In this particular case, the model filled in the gaps by generating a list of ten jobs, starting with \enquote{Volunteer, Counseling Center, 2006 - 2007} and ending with \enquote{Researcher at Florida State University, present - current position.} The model made these jobs seem more believable by using information from the education and publication histories to create a credible sequence. 

Of all the RAG-related response errors, we find the potential of missing context the most concerning. That one of the most recent OpenAI GPT models\footnote{gpt-3.5-turbo-16k-0613} could construct what is essentially a \textit{more credible} hallucination is an issue that deserves further study.

\subsection{Unusual Formatting - 4 Responses (19.0\%)}

Several of the CVs uploaded were substantially different in their formatting and content. In the case of subject 58, the CV was particularly long with detailed descriptions of tasks and responsibilities. In particular, the date information for each job listing followed these long and detailed descriptions, which placed date information from the \textit{previous} job in close proximity to the position and institution of the \textit{subsequent} job. The model associated the two disjoint items into a single element in the list. Other examples followed this pattern. If the model encountered context that was too dissimilar to common CV structure -- particularly with respect to dates, the responses would be subtly wrong.

As with \textit{Context Drift}, we would consider this behavior to be \textit{hallucination-adjacent}, since it is inappropriate information delivered in a credible way.

\subsection{Incomplete Context - 1 Response (4.8\%)}
In one case, subject 52 provided a DOI as a pointer to the information about their (single) publication. The model we used did not have the capability to search online data, and as a result, had to rely on the context. Using the context, the model produced what the user considered an incomplete response that did not specify anything beyond what was provided in the context. This points to the fact that many, if not most people consider LLMs to be a kind of information retrieval system (e.g. the legal cases discussed in Section~\ref{sec:introduction}), 
and so they feel that the LLM system \textit{should} be able to determine the journal from the DOI. 

\section{Discussion}
\label{sec:discussion}

Over the centuries, our understanding of information retrieval has changed. We used to consult elders and oracles, who would provide information that would often require subsequent \enquote{interpretation.} The development of libraries led to cataloging and indexing. Computers and high-speed storage begat a lineage of information retrieval systems that led from Gerard Salton's SMART document retrieval system~\cite{salton1965smart} to Google's PageRank~\cite{page1998pagerank}. All these systems used  statistical, syntactic, and semantic procedures for the analysis of information and the identification of relevant items. Precision and Recall are the currency of these systems.

Large Language Models are more like oracles than libraries. A `completion' LLM is restricted to the data they have been trained on and the information in the prompt. As we have seen, subject to a substantial risk to the generation of credible-sounding but false statements, or as we have come to call them, hallucinations. They are most accurate when they can perform in domains where the training set has many correct, largely aligned, examples. A model with many conflicting sources or with poisoned data will struggle to produce nonfiction. Since answers are synthesized across the training set, precision and recall are rendered useless. Instead, like the oracles of old, these results must be interpreted, often by other LLMs through benchmarks such as HellaSwag~\cite{zellers2019hellaswag}, TruthfulQA~\cite{lin2021truthfulqa}, and MMLU~\cite{hendrycks2020measuring}.

Retrieval-augmented generation is a hybrid oracle-library. RAGs dynamically retrieve and incorporate data from an external sources based on the user query. The query is combined with the search results, and an answer is generated from that. This has proven to be an effective approach in reducing hallucinations. Since sources for context can be identified, once more search matters. But LLMs can add their own biases and distort the response. 

Our research has shown that in the presence of complex or misleading search results, a RAG system may often get things wrong. A missing section may lead to hallucinations, an unconventional placement of dates may result in the time from one event being attributed to another. 

Or, if the training data in the model is sufficiently biased, the \textit{generated} answer may be substantially different from the ground truth. When a Chinese spy balloon drifted over the western US in February 2023, the news was filled with reports. But when checkers at the  the GDELT project\footnote{Personal communication with Kalev Leetaru of \surl{https://www.gdeltproject.org/}} went over LLM translations and summaries of these reports, they found that the models had often changed the balloon to something more high-tech, like a hypersonic spy plane. News reporting on China has so long emphasized its technological prowess that the models were unable to handle something as mundane as a balloon. The oracle had added its perspective. 

In our experiments, the use of context prompting, such as used in RAG systems, significantly improved accuracy levels to 94\%, compared to the un-augmented model, which had an accuracy of only 7.3\%. While this demonstrates a substantial improvement, the RAG system may still struggle to provide accurate information to users in cases where the context provided falls beyond the scope of the model's training data. The scale of this issue becomes more apparent when considering the estimated 8.5 billion Google searches conducted daily in 2023.\footnote{\surl{https://www.demandsage.com/google-search-statistics}} An equivalent RAG system could potentially generate misleading or incorrect information in approximately 500 million results per day. Furthermore, in real-world conditions, where the retrieval step may introduce additional error, the number of misleading or false responses should be much higher. 

In the view of the author, these errors, and the harm they may cause, cannot be wiped away by having a disclaimer at the bottom of the page that says, as with Google's Bard: \enquote{may display inaccurate info, including about people, so double-check its responses.}

\section{Conclusions}
\label{sec:conclusions}

This study  demonstrates the significant role of context in enhancing the accuracy of responses generated by language model invocations with retrieval-augmented generation (RAG). The incorporation of context resulted in a remarkable 18-fold improvement in correctly navigating the text, with an impressive accuracy rate of 94\%. This represents an 18-fold improvement over situations where context is absent, emphasizing the critical importance of context in LLM responses where accuracy is important.

Our analysis of the small percentage (6.04\%) of incorrect responses has yielded insights into the types of errors that can occur despite providing accurate context. We identified five categories of error, with \textit{noisy context} being the most prevalent, followed by \textit{mismatched instructions and context}, \textit{context-based synthesis}, \textit{unusual formatting}, and \textit{incomplete context}. Each of these error types manifests differently, but they share a commonality: they all distort the output in various ways that are often believable, which can be particularly dangerous when not closely scrutinized.

Our findings raise important issues for the future of RAG systems. Particularly alarmingly is the capability of models to generate credible hallucinations by interpolating between factual content. This represents a danger that demands further research attention. Furthermore, our results highlight the implications of unusual formatting and incomplete context on the reliability of RAG responses.

In the instance where participant 52 expected a DOI to lead to a comprehensive paper description, it suggests that the general public holds certain beliefs regarding the capabilities of LLMs. This presents an interesting opportunity for future research, focusing on managing user expectations and enhancing prompt engineering to optimize RAG systems.

In conclusion, our study validates the effectiveness of context in improving the quality of responses from RAG systems. Nevertheless, it also exposes the nuances of error types that arise even with accurate context, suggesting a path forward for enhancing RAG systems through improved understanding of context utilization and prompt engineering. These findings contribute to advancing the field and lay the groundwork for the next generation of more reliable, context-aware machine learning applications.



\newpage
\begin{thebibliography}{10}

\bibitem{alkaissi2023artificial}
Hussam Alkaissi and Samy~I McFarlane.
\newblock Artificial hallucinations in chatgpt: implications in scientific
  writing.
\newblock {\em Cureus}, 15(2), 2023.

\bibitem{bender2020climbing}
Emily~M Bender and Alexander Koller.
\newblock Climbing towards nlu: On meaning, form, and understanding in the age
  of data.
\newblock In {\em Proceedings of the 58th annual meeting of the association for
  computational linguistics}, pages 5185--5198, 2020.

\bibitem{borji2023categorical}
Ali Borji.
\newblock A categorical archive of chatgpt failures.
\newblock {\em arXiv preprint arXiv:2302.03494}, 2023.

\bibitem{chen2023benchmarking}
Jiawei Chen, Hongyu Lin, Xianpei Han, and Le~Sun.
\newblock Benchmarking large language models in retrieval-augmented generation.
\newblock {\em arXiv preprint arXiv:2309.01431}, 2023.

\bibitem{feldman2023trapping}
Philip Feldman, James~R Foulds, and Shimei Pan.
\newblock Trapping llm hallucinations using tagged context prompts.
\newblock {\em arXiv preprint arXiv:2306.06085}, 2023.

\bibitem{hendrycks2020measuring}
Dan Hendrycks, Collin Burns, Steven Basart, Andy Zou, Mantas Mazeika, Dawn
  Song, and Jacob Steinhardt.
\newblock Measuring massive multitask language understanding.
\newblock {\em arXiv preprint arXiv:2009.03300}, 2020.

\bibitem{heyman2023impact}
Tom Heyman and Geert Heyman.
\newblock The impact of chatgpt on human data collection: A case study
  involving typicality norming data.
\newblock {\em Behavior Research Methods}, pages 1--8, 2023.

\bibitem{lewis2020retrieval}
Patrick Lewis, Ethan Perez, Aleksandra Piktus, Fabio Petroni, Vladimir
  Karpukhin, Naman Goyal, Heinrich K{\"u}ttler, Mike Lewis, Wen-tau Yih, Tim
  Rockt{\"a}schel, et~al.
\newblock Retrieval-augmented generation for knowledge-intensive nlp tasks.
\newblock {\em Advances in Neural Information Processing Systems},
  33:9459--9474, 2020.

\bibitem{lin2021truthfulqa}
Stephanie Lin, Jacob Hilton, and Owain Evans.
\newblock Truthfulqa: Measuring how models mimic human falsehoods.
\newblock {\em arXiv preprint arXiv:2109.07958}, 2021.

\bibitem{manakul2023selfcheckgpt}
Potsawee Manakul, Adian Liusie, and Mark~JF Gales.
\newblock Selfcheckgpt: Zero-resource black-box hallucination detection for
  generative large language models.
\newblock {\em arXiv preprint arXiv:2303.08896}, 2023.

\bibitem{mckenna2023sources}
Nick McKenna, Tianyi Li, Liang Cheng, Mohammad~Javad Hosseini, Mark Johnson,
  and Mark Steedman.
\newblock Sources of hallucination by large language models on inference tasks.
\newblock {\em arXiv preprint arXiv:2305.14552}, 2023.

\bibitem{page1998pagerank}
Lawrence Page, Sergey Brin, Rajeev Motwani, and Terry Winograd.
\newblock The pagerank citation ranking: Bring order to the web.
\newblock Technical report, Technical report, stanford University, 1998.

\bibitem{salton1965smart}
Gerard Salton and Michael~E Lesk.
\newblock The smart automatic document retrieval systems—an illustration.
\newblock {\em Communications of the ACM}, 8(6):391--398, 1965.

\bibitem{shen2023large}
Tao Shen, Guodong Long, Xiubo Geng, Chongyang Tao, Tianyi Zhou, and Daxin
  Jiang.
\newblock Large language models are strong zero-shot retriever, 2023.

\bibitem{simon1996sciences}
Herbert~A Simon.
\newblock {\em The sciences of the artificial}.
\newblock MIT press, 1996.

\bibitem{teixeira2020curriculum}
Jaime~A Teixeira~da Silva, Judit Dobr{\'a}nszki, Aceil Al-Khatib, and
  Panagiotis Tsigaris.
\newblock Curriculum vitae: challenges and potential solutions.
\newblock {\em KOME: An International Journal of Pure Communication Inquiry},
  8(2):109--127, 2020.

\bibitem{veselovsky2023prevalence}
Veniamin Veselovsky, Manoel~Horta Ribeiro, Philip Cozzolino, Andrew Gordon,
  David Rothschild, and Robert West.
\newblock Prevalence and prevention of large language model use in crowd work,
  2023.

\bibitem{wong2023security}
Ann~Yi Wong, Eyasu~Getahun Chekole, Mart{\'\i}n Ochoa, and Jianying Zhou.
\newblock On the security of containers: Threat modeling, attack analysis, and
  mitigation strategies.
\newblock {\em Computers \& Security}, 128:103140, 2023.

\bibitem{zellers2019hellaswag}
Rowan Zellers, Ari Holtzman, Yonatan Bisk, Ali Farhadi, and Yejin Choi.
\newblock Hellaswag: Can a machine really finish your sentence?
\newblock {\em arXiv preprint arXiv:1905.07830}, 2019.

\bibitem{zhuo2023exploring}
Terry~Yue Zhuo, Yujin Huang, Chunyang Chen, and Zhenchang Xing.
\newblock Exploring {AI} ethics of {ChatGPT}: A diagnostic analysis.
\newblock {\em arXiv preprint arXiv:2301.12867}, 2023.

\end{thebibliography}

\end{document}